\documentclass[nohyperref]{article}

\usepackage{microtype}
\usepackage{graphicx}
\usepackage{subfigure}
\usepackage{booktabs} 

\usepackage{hyperref}

\usepackage{algpseudocode}
\algnewcommand\algorithmicforeach{\textbf{for every}}
\algdef{S}[FOR]{ForEach}[1]{\algorithmicforeach\ #1\ \algorithmicdo}

\usepackage[accepted]{icml2022}

\usepackage{amsmath}
\usepackage{amssymb}
\usepackage{mathtools}
\usepackage{amsthm}
\usepackage{bbm}

\usepackage{caption}
\usepackage{pifont} 
\usepackage{algorithm,algpseudocode} 
\usepackage{multicol,multirow}
\usepackage{makecell}
\usepackage{hhline}
\usepackage{colortbl}
\usepackage{listings}
\usepackage{mdframed}
\usepackage[normalem]{ulem}
\usepackage{xcolor}
\usepackage{tikz}
\usepackage{enumitem}

\usepackage[capitalize,noabbrev]{cleveref}

\theoremstyle{plain}

\theoremstyle{definition}

\theoremstyle{remark}

\newcommand{\todoc}[2]{{\textcolor{#1}{\textbf{#2}}}}
\newcommand{\todored}[1]{{\todoc{red}{\textbf{[#1]}}}}

\DeclareMathOperator*{\argmin}{arg\,min}

\usepackage[textsize=tiny]{todonotes}

\newcommand{\xz}[1]{\todored{XZ: #1}}

\begin{document}

\twocolumn[

\icmltitle{Constrained Optimization with Dynamic Bound-scaling for Effective NLP Backdoor Defense}



\icmlsetsymbol{equal}{*}

\begin{icmlauthorlist}
\icmlauthor{Guangyu Shen}{equal,yyy}
\icmlauthor{Yingqi Liu}{equal,yyy}
\icmlauthor{Guanhong Tao}{yyy}
\icmlauthor{Qiuling Xu}{yyy}
\icmlauthor{Zhuo Zhang}{yyy}
\icmlauthor{Shengwei An}{yyy}
\icmlauthor{Shiqing Ma}{comp}
\icmlauthor{Xiangyu Zhang}{yyy}
\end{icmlauthorlist}

\icmlaffiliation{yyy}{Department of Computer Science, Purdue University, West Lafayette, IN, USA}
\icmlaffiliation{comp}{Department of Computer Science, Rutgers University, Piscataway, NJ, USA}

\icmlcorrespondingauthor{Guangyu Shen}{shen447@purdue.edu}
\icmlcorrespondingauthor{Yingqi Liu}{liu1751@purdue.edu}

\icmlkeywords{Machine Learning, ICML}

\vskip 0.3in
]



\printAffiliationsAndNotice{\icmlEqualContribution} 

\begin{abstract}
We develop a novel optimization
method for NLP backdoor inversion.
We leverage a dynamically reducing temperature coefficient in the softmax function to provide changing loss landscapes to the optimizer such that the process gradually focuses on the ground truth trigger, which is denoted as a one-hot value in a convex hull. 
Our method also features a temperature rollback mechanism to step away from local optimals, exploiting the observation that  local optimals can be easily determined in NLP trigger inversion (while not in general optimization). 
We evaluate the technique on over 1600 models (with roughly half of them having injected backdoors) on 3 prevailing NLP tasks, with 4 different backdoor attacks and 7 architectures. Our results show that the technique is able to effectively and efficiently detect and remove backdoors, outperforming 4 baseline methods.
\end{abstract}

\section{Introduction}
Backdoor attack~\cite{gu2017badnets, liu2017trojaning} poses a severe threat to the state-of-the-art NLP models~\cite{devlin2018bert,vaswani2017attention,radford2019language}.
By trojaning a model, attackers can induce model misclassification with some pre-defined token sequence (i.e., trigger). Some recent attacks can even use specific sentence structure and para-phrasing behavior as the trigger~\cite{qi2021hidden,qi2021turn}, making the attacks very stealthy. 
Hence, detecting backdoor in a pre-trained model and removing such backdoor are critical for secure applications of these models. 
Trigger inversion is an effective approach for scanning model backdoors in the vision domain. It leverages gradient descent to derive a trigger. 
However, these approaches cannot be simply extended to NLP models due to the discrete nature of these models. Specifically, the input space of NLP models are words or tokens that are sparse in the input space. A naive trigger inversion algorithm may yield input values that do not correspond to any valid words or tokens. 

There are a number of pioneering works in defending NLP model backdoors,
for example, identifying and removing triggers from input sentences via grammar checking~\cite{pruthi2019combating}, language model~\cite{qi2020onion} and model internal analysis~\cite{chen2021mitigating}.  
More can be found in Section~\ref{related work}.
These techniques have encouraging results in their targeted scenarios. However, some have certain assumptions such that they may not be as effective when the assumptions are not satisfied, for example, when the trigger contains multiple tokens.

\begin{figure}[t!]
\centering
    \includegraphics[width=\columnwidth]{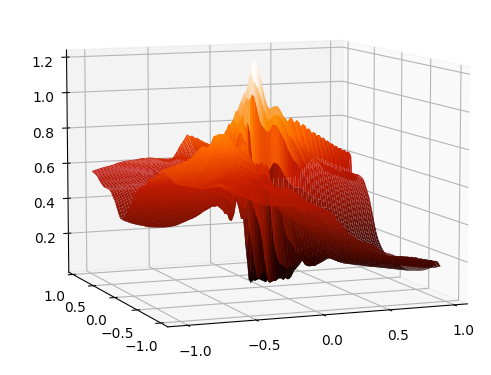}
    
\caption{Difficult loss landscape with a low temperature}
\label{fig:fig1}
\end{figure}

In this paper, we develop a novel optimization
method for general NLP backdoor inversion. Instead of directly 
inverting word/token values, we define a convex hull over all tokens. A value in the hull is a weighted sum of all the token values in the dictionary. The weight vector hence has a dimension number equals to the number of tokens in the dictionary (e.g., 30522 in BERT) and the sum of all dimensions equals to 1.
We then aim to invert trigger weight vector(s). Ideally, an inverted weight vector should have a one-hot value, indicating a token in the dictionary. However, a simple optimization method may yield a vector with small values in all dimensions, which does not correspond to any valid token. 
We leverage the {\em temperature} coefficient in the $\mathit{softmax}$ function to control the optimization result. According to~\cite{hinton2015distilling,jang2016categorical,guo2017calibration}, a low temperature produces more confident result, meaning the result weight vector tends to have only one large dimension and the rest very small (similar to a one-hot value).
However, directly using a low temperature in optimization renders a very rugged loss landscape, causing the optimizer stuck in local optimals, namely, producing one-hot values corresponding to valid tokens that do not achieve high {\em attack success rate} (ASR).
In Figure~\ref{fig:fig1}, we render the loss landscape for a real-world trojaned model from~\cite{trojai} when a low temperature is used. Observe that there are many steep peaks, making optimization very challenging. 
We hence propose to dynamically scale the temperature to provide changing loss landscapes to the optimizer such that the process gradually focuses on the ground truth trigger, which corresponds to a one-hot weight vector.
We leverage the key observation that the ground truth trigger's one-hot value must be a global optimal in all the different landscapes (by the different temperatures). Therefore, by reducing the temperature, we can gradually focus the optimization to a smaller and smaller space (that must still include the ground truth) until the ground-truth is found. 
Our method also features a temperature rollback mechanism to escape from local optimals. Specially, if using a low temperature continues to produce local optimals, it steps back and uses a higher temperature. It exploits another key observation that local optimals can be easily determined in NLP trigger inversion (while not in general optimization). Specifically, the ground truth value, i.e., the global optimal, must have a very small inversion loss.  Hence, inverted values that do not have a small inversion loss must be local optimals. 

We evaluate our technique on backdoor detection using 1584 transformer models from TrojAI~\cite{trojai} rounds 6-8 datasets and 120 models from 3 advanced stealthy NLP backdoor attacks.
Our method consistently outperforms 4 stat-of-the-art baselines, ASCC~\cite{dong2021towards}, UAT~\cite{wallace2019universal}, T-Miner~\cite{azizi2021t} and Genetic Algorithm~\cite{alzantot2018generating}, having up to $23.0\%$, $12.2\%$, $43.4\%$ and $27.8\%$ higher detection accuracy, respectively. 
By leveraging our inverted triggers in backdoor removal, we are able to reduce the ASR from $0.987$ to $0.058$. We will release the code upon publication. 

\vspace{-5pt}
\section{Related Work}
\label{related work}

A large body of works have been proposed for backdoor attack and defense. Early works mainly focus on the computer vision domain~\cite{liu2017trojaning,gu2017badnets,chen2017targeted,liu2020reflection,salem2020dynamic,wang2019neural,liu2019abs,shen2021backdoor,xu2019detecting}. Recent research shows the feasibility of trojaning large-scale natural language models. For example, BadNL~\cite{chen2021badnl} proposes to use character, word or sentence as triggers in NLP domain. 
\textit{Hidden Killer Attack}~\cite{qi2021hidden} uses a specific sentence structure as the backdoor trigger.
SOS~\cite{yang2021rethinking} proposes to improve the stealthiness of backdoor attacks by adding the sub-strings of the trigger phrase with correct labels to the training set and forcing the model to learn the long phrase as the true trigger.
\textit{Combination Lock Attack}~\cite{qi2021turn} and Hidden backdoor~\cite{li2021hidden} train a paraphrasing model and use this paraphrasing model as the trigger.
There is another line of works that focuses on poisoning pre-trained models in NLP domain~\cite{kurita2020weight, shen2021backdoorpre, zhang2021red}. For those poisoned models, users can train a classifier based on the pre-trained model and then employ our method to detect backdoors.

Backdoor detection for NLP models is a new area. It falls into two categories. 
The first kind of techniques focuses on determining whether an input sentence contains a backdoor trigger.
Onion~\cite{qi2020onion} uses the perplexity of input sentences to detect backdoor inputs.
Pruthi et al.~\cite{pruthi2019combating} propose to use word checker to remove character triggers in input sentences.
BKI~\cite{dai2019backdoor} leverages the word impact in LSTM models for backdoor sample detection.
The second kind aims to determine whether a NLP model contains backdoors without the access to sentences with triggers.
Our method falls in this category.
MNTD~\cite{xu2019detecting} utilizes meta-learning to train a large set of shadow models for backdoor detection. MNTD mainly targets models in the computer vision domain and was only evaluated on one-layer LSTM models in NLP domain.
T-miner~\cite{azizi2021t} trains a sequence-to-sequence generative model to reverse-engineer backdoor triggers. It then uses the attack success rate of generated triggers to determine whether a model contains backdoors. T-miner uses random samples to train the generative model and does not require benign sentences.
A few adversarial example generation methods in NLP domain~\cite{alzantot2018generating,wallace2019universal,dong2021towards} aim to generate a small perturbation to induce misclassification. They can be adapted to reverse-engineer triggers for backdoor detection. We evaluate on two state-of-the-art adversarial generation methods, ASCC~\cite{dong2021towards} and Genetic Algorithm (GA)~\cite{alzantot2018generating}, in Section~\ref{sec:eval}.

\vspace{-5pt}
\section{Preliminaries}
In this section, we formally define NLP backdoor attack and defense.

\subsection{NLP Backdoor Attack}

For the notation simplicity, we consider a textual classification task. Given a transformer model $f(\theta)$ parameterized by $\theta$ and a clean dataset 
$\mathcal{D} = \{\mathcal{X},\mathcal{Y}\}$, where $x = \{x_i\}_{i=1}^n \in \mathcal{X}$ is a text sequence with $n$ tokens, $y \in \mathcal{Y}$ is the corresponding label,
An NLP backdoor attack aims to train a model $f(\theta^*)$ associated with a secret $m$ tokens sequence  ${t^*} = \{t^*_i\}_{i=1}^m$ (i.e., the {\em trigger}) and a target label $y^* \in \mathcal{Y}$. It
first constructs a poisoned dataset $\mathcal{D}_p = \{\mathcal{D} \cup \mathcal{D}^* \}$, 
where $\mathcal{D}^* = \{\mathcal{X}^*,y^*\}$, $x^* = \{x_i\}_{i=1}^n \oplus \{t^*_i\}_{i=1}^m \in \mathcal{X^*}$, and then minimizes the following training loss.
\setlength\abovedisplayskip{3pt}
\setlength\belowdisplayskip{2pt}
\begin{equation}
\label{backdoor loss def}
\begin{aligned}
\small
\mathcal{L}_{\mathcal{D}_p}(\theta^*) &= \mathop{\mathbb{E}}_{(x,y)\sim  \mathcal{D}} \mathcal{L} (f(x;\theta^*),y) \\
 &+ \mathop{\mathbb{E}}_{(x^*,y^*) \sim  \mathcal{D}^*} \mathcal{L}(f(x^*;\theta^*),y^*)
\end{aligned}
\end{equation}
where $l(\cdot,\cdot)$ is the cross entropy loss and $\oplus$ denotes {\em the trigger injection} operation, which could be insertion and replacement at various positions~\cite{chen2021badnl,dai2019backdoor}. The attackers usually set $m \ll n$ for the purpose of stealthiness~\cite{kurita2020weight,yang2021careful}. 

\subsection{NLP Backdoor Defense}

\textbf{Threat Model}. Similar to \cite{azizi2021t,shen2021backdoor,liu2019abs,wang2019neural}, we assume that the defender gains the full access to  a subject model $f(\theta)$, which may be trojaned or clean, and a small subset of the clean dataset $\mathcal{D}^\prime \in \mathcal{D}$. 
 The defender has no knowledge of the injected trigger $t^*$, target label $y^*$ or the poisoned dataset $\mathcal{D}^*$, if the model is trojaned. We consider two types of defense. The first is to determine if the subject model contains any backdoor. The second is to remove the backdoor if there is one. Therefore, runtime backdoor sample detection~\cite{qi2020onion,9508422,pruthi2019combating} that aims to determine if an input sample contains any backdoor trigger is beyond our scope (as it requires poisoned samples).

\subsubsection{Trigger Inversion}
Trigger inversion~\cite{liu2019abs,wang2019neural,shen2021backdoor,wang2020practical} aims to reverse engineer the injected trigger and the corresponding target label through optimization. Specifically, given a subject model, it treats each label  a potential target label (of an attack) and tries to derive a token sequence which is able to flip all samples to the target class. Since the procedure may identify a number of such trigger and target label pairs, it usually reports the most prominent pair  based on certain metrics (e.g., the smallest trigger with the highest ASR).
Mathematically, for each label $y_i \in \mathcal{Y}$, it tries to find a $t^{[y_i]}$ to minimize the loss:
\begin{equation}
\label{backddoor trigger inversion def}
\begin{aligned}
\small
\mathcal{L}_{inv}(t^{[y_i]},y_i,\theta^*) = \mathop{\mathbb{E}}_{x \sim  \mathcal{D}^\prime}\mathcal{L}(f (x\oplus t^{[y_i]};\theta^*),y_i)
\end{aligned}
\end{equation}
Observe that by iterating  over all possible labels, the above procedure always produces a set of inversed triggers, one for each label. To determine if a model has any backdoor, we have the following assumption.

\textbf{Assumption \uppercase\expandafter{\romannumeral1}:} Given a trojaned model $f(\theta^*)$ with target label $y^*$, $\mathcal{L}_{inv}(t^{[y_i]},y_i,\theta^*) \ll \mathcal{L}_{inv}(t^{[y_j]},y_j,\theta^*),  \forall y_j \neq y_i \in \mathcal{Y}\ $ if $y_i = y^*$. 

Intuitively, it is easier to flip samples to the ground-truth target label than to other labels for a trojaned model. We use $(\hat{t},\hat{y}) = \argmin\mathcal{L}_{inv}(t^{[y_i]},y_i,\theta^*), \forall y_i \in \mathcal{Y}$ to denote the \textbf{optimal trigger estimation} for a model $f$. 




\subsubsection{Backdoor Detection}
Backdoor detection aims to determine if a given NLP model has a backdoor. It can be defined as a binary classification task over a set of benign and trojaned models. In particular, given a model set $\{(f(\theta_k),\mathcal{D}^\prime_k,\omega_k)\}_{k=1}^N$, a backdoor detection algorithm tries to derive a binary function $\mathcal{F}$ so that
\begin{equation}
\label{backdoor detection def}
\begin{aligned}
\small
\mathcal{F}(\theta_k,\mathcal{D}_k^\prime) = \omega_k, \forall{k} \in N
\end{aligned}
\end{equation}
where $\omega_k \in \{0,1\}$ 
denotes the benignity of model $k$. Here, we have an assumption regarding models. 

\textbf{Assumption \uppercase\expandafter{\romannumeral2}:} Assume $(\hat{t}_t,\hat{y}_t)$ and $(\hat{t}_b,\hat{y}_b)$ denote the optimal trigger estimations for a trojaned model $f_t$ and a benign model $f_b$, respectively, we have 
$\mathcal{L}_{inv}(\hat{t}_t,\hat{y}_t,\theta_t) \ll \mathcal{L}_{inv}(\hat{t}_b,\hat{y}_b,\theta_b)$.

Intuitively, due to the existence of backdoor, it is easier to find a token sequence that can flip a trojaned model's prediction than a benign model. Based on the assumption, a straightforward way to realize the detection function is the following.
\begin{equation}
\label{backdoor detection def rewrite}
\small
\mathcal{F}(\theta_k,\mathcal{D}_k^\prime) = \left \{
\begin{aligned}
&1,  &\mathcal{L}_{inv}(\hat{t}_k,\hat{y}_k,\theta_k) < \beta  & \\
&0,  &\mathcal{L}_{inv}(\hat{t}_k,\hat{y}_k,\theta_k) > \beta \\
\end{aligned}\right.
\end{equation}
where $\beta$ is a learnable threshold to distinguish benign and trojaned models.
Note that more complex methods to construct the detection function are possible such as supervised training based on logits and/or model internals. 
In this paper, we find that the simple threshold based method is sufficient. 

\subsubsection{Backdoor Removal}
Backdoor removal aims to eliminate an identified backdoor without hurting the model's clean accuracy. A standard way is to perform model unlearning\cite{wang2019neural}. Particularly, it optimizes Eq.\ref{backdoor loss def} inversely as follows.  
\begin{equation}
\label{backdoor removal loss def}
\begin{aligned}
\small
\argmin_{\theta^*}&[\mathop{\mathbb{E}}_{(x,y)\sim  \mathcal{D}} \mathcal{L} (f(x;\theta^*),y)  \\
&- \mathop{\mathbb{E}}_{(x^*,y^*) \sim  \mathcal{D}^*}\mathcal{L}(f(x^*;\theta^*),y^*)]
\end{aligned}
\end{equation}
In practice,  we approximate the unknown $(t^*,y^*)$ by the inversion results  $(\hat{t},\hat{y})$, and Eq.\ref{backdoor removal loss def} can be rewritten as 
\begin{equation}
\label{backdoor removal loss def rewrite}
\begin{aligned}
\small
\argmin_{\theta^*}\mathop{\mathbb{E}}_{(x,y)\sim  \mathcal{D}^\prime}[\mathcal{L} (f(x;\theta^*),y) - \mathcal{L}(f(x\oplus \hat{t};\theta^*),\hat{y})]
\end{aligned}
\end{equation}
Therefore, the accurate estimation of $(\hat{t},\hat{y})$ is crucial for both  backdoor removal and detection tasks.

\vspace{-5pt}
\section{Methodology}

\begin{figure*}[t]

\end{figure*}

\begin{figure*}[t!]
    \centering
    \subfigure[High temperature]{
        \includegraphics[width=2in]{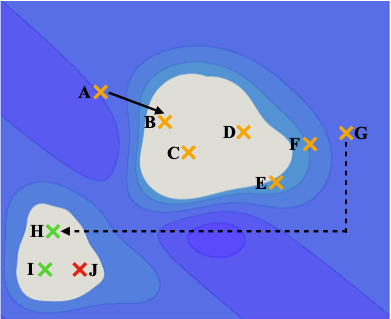}
    }
    \subfigure[Moderate temperature]{
	\includegraphics[width=2in]{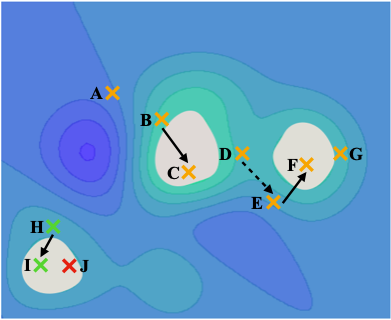}
    }
    \subfigure[Low temperature]{
	\includegraphics[width=2in]{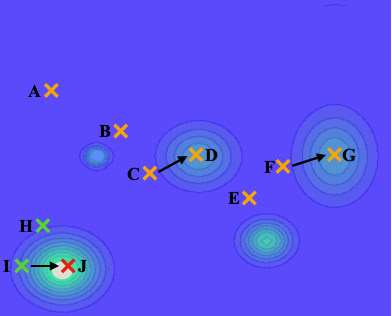}
    }
    \caption{Illustration of Temperature Scaling and Backtracking in Different Loss Landscapes 
    }
    \label{illustration}

\label{fig:illusration}
\vspace{-15pt}
\end{figure*}

In this section, we first introduce the challenges in applying trigger inversion for backdoor defense and then
present our method.

As a standard procedure, modern transformer models have an input embedding function $e(\cdot)$ to map an input token sequence to a lower dimensional embedding vector, and then feed the embeddings forward through the network.
The trigger inversion process can be revised as follows with $e(x)$ the embeddings of an input $x$. 
\begin{equation}
\label{backddoor trigger inversion def rewrite}
\begin{aligned}
\small
\argmin_{t}\mathcal{L}_{inv} = \mathop{\mathbb{E}}_{x \sim  \mathcal{D}}\mathcal{L}(f(e(x),e(t);\theta),y)
\end{aligned}
\end{equation}
Note that Eq.\ref{backddoor trigger inversion def rewrite} is not end-to-end differentiable for two reasons. First, the optimization variable $t$ is defined on a discrete space $N_+^{m \times p}$, where $m$ is the trigger length and $p$ is the entire dictionary size (e.g., 30522 for BERT). Second, gradient will be cut while propagating through the input embedding layer since it is a table lookup operation~\cite{turian-etal-2010-word,peters2017semi,gardner2018allennlp}.


\subsection{Defining A Convex Hull for the Input Space}
In order to address the non-differentiability issue, we perform a linear relaxation of the discrete input space by defining a convex hull. 

\textbf{Proposition \uppercase\expandafter{\romannumeral1}:} Let $\mathcal{C} = \{c_1,c_2,\cdots,c_p\}$ be the set of all possible tokens in dictionary. The convex hull over input space can be defined as $\mathcal{V} = \{\sum_{j=1}^p \alpha_j e(c_j) \mid  \sum_{j=1}^p \alpha_j = 1, \alpha_j \geq 0 \}$

Therefore, an arbitrary input token $t_i$ can be represented in the convex hull: $\mathcal{V}(t_i) = \sum_{j=1}^p \alpha_{ij} e(c_j)$ where $ \sum_{j=1}^p \alpha_{ij} = 1, \alpha_{ij} \geq 0$. Similar to \cite{dong2021towards}, by introducing $w_{ij} \in \mathcal{R}^{m\times p}$ and leveraging $\mathit{softmax}$, 
we can satisfy the coefficient constraint by the following transformation.
\begin{equation}
\label{Softmax}
\begin{aligned}
\small
\alpha_{ij} = \frac{\exp{(w_{ij})}}{\sum_{j=1}^p \exp{(w_{ij})}}
\end{aligned}
\end{equation}
After defining the convex hull over input space, we can avoid directly optimizing tokens in the trigger sequence, but rather optimizing the coefficients $w_{ij}$.
For example, backdoor detection is to solve the following optimization problem.
\begin{equation}
\label{backddoor trigger inversion def relax}
\begin{aligned}
\small
\argmin_{\alpha_{ij}}\mathop{\mathbb{E}}_{x \sim  \mathcal{D}}\mathcal{L} (f(e(x),\{\sum_{j=1}^p \alpha_{ij} e(c_j)\}_{i=1}^m;\theta),y)
\end{aligned}
\end{equation}
In the NLP adversarial example generation technique ASCC~\cite{dong2021towards}, for each token in the input, they have a small substitute token list, and then define a convex hull over such list. The optimization hence looks for a value in the convex hull (a linear combination of the substitute tokens) that can induced misclassification. The substitute list is small for each token (in the scale of 10). 
In contrast, we consider all possible tokens (more than 30000) in our convex hull as the trigger can be any token and all tokens  share the same convex hull. 
This makes optimization substantially more challenging and motivates our unique design. 





\subsection{Gradual Focusing with Temperature Scaling and Backtracking}

Directly solving Eq.~\ref{backddoor trigger inversion def relax} 
produces excessive false results,
namely, unrealistic vectors (in the hull) that do not correspond to any legitimate tokens in the dictionary. 
For example, it may yield a vector with all the $\alpha_{ij}$ having some small values. While the vector can incur a very small loss value, it does not constitute any valid token.
This is because NLP models are only trained on a limited set of vectors (from valid tokens) and their behaviors in the whole input space are largely undefined. While this is not a problem for forward computation, it makes inversion extremely difficult.

Figure~\ref{fig:illusration} (a) shows an example contour of loss landscape when we directly optimize Eq.~\ref{backddoor trigger inversion def relax}.  
A point denotes an $\alpha$ vector value (after dimension reduction), and its color denotes the loss value for the point.
The brighter the color, the smaller the loss. Therefore, the white points denote global optimals. The points brighter than their surroundings and do not have a white color denote local optimals. 

Here, point $J$
corresponds to the ground truth trigger token. Note that its $\alpha$ vector is a one-hot value, with the dimension corresponding to the token having value 1 and the rest 0. Due to the undefined behaviors, all the points in a large surrounding area of $J$ also yield a comparable small loss. As such, the optimization produces many false results.    

Ideally, we would want the optimization to produce a one-hot value. This can be achieved by controlling the {\em temperature} in the $\mathit{softmax}$ function.
\begin{equation}
\label{Softmax with t}
\begin{aligned}
\small
\alpha_{ij} = \frac{\exp{(w_{ij}}/{\lambda})}{\sum_{j=1}^p \exp{(w_{ij}}/{\lambda})}
\end{aligned}
\end{equation}
Specifically, coefficient $\lambda$ in Eq.~\ref{Softmax with t} controls result confidence,  with a default value 1. 
When it is large ($\lambda\gg 1$), called {\em high temperature},  all the dimensions in the optimization result tend to have small differences, indicating low confidence. When it is small ($\lambda\ll 1$), called {\em low temperature}, one of the dimentions tends to substantially stand out (similar to a one-hot value). Intuitively, this is because 
a low temperature enlarges the numerical difference among dimensions and after going through $\mathit{softmax}$, such difference is further magnified due to its exponential nature. As such, the value of top-1 dimension becomes exponentially larger than the others.
Therefore, a naive idea is to use a very low temperature in optimization.
However, this is problematic because the loss landscape becomes rougher with a lower temperature and hence the optimization tends to produce local optimals, e.g., one-hot values whose loss values are large, meaning valid tokens failing to incur high ASR.

Figure~\ref{fig:illusration} (c) shows the loss landscape when we use a low temperature. Observe that while $J$ remains a global optimal, there are many other optimals, including global and local. They largely correspond to valid tokens but most of them do not have a small loss value. Direct optimization in such a landscape is not effective either.

\smallskip
\noindent
{\bf Our Solution.}
We make a key observation that the {\it ground truth} (GT) token remains a global optimal for all possible temperatures (e.g., point $J$). When the temperature is high, the landscape is smooth and a large surrounding area of the GT point has a small loss value. We call it the {\em optimal zone} (OZ). With temperature decreasing, OZ also decreases while still including the GT.  
Given a specific temperature, the optimization may yield any point in the corresponding OZ. 

Our overarching idea is hence the following. We start with a high temperature (and hence a smooth landscape and a large OZ), e.g., Figure~\ref{fig:illusration} (a). 
Assume the optimization yields some point in the OZ, e.g., $H$ in Figure~\ref{fig:illusration} (a). 
We then lower the temperature and resume optimization from $H$, which essentially means that we are using a landscape that is more focused and has a smaller OZ, e.g., Figure~\ref{fig:illusration} (b).
This time, the optimization yields a point in the smaller OZ, e.g., $I$ in (b). 
According to our observation, all the OZ's include the GT. By gradually decreasing the temperature, we guide the optimization to become more and more focused, until the GT is found, i.e., point $J$ in Figure~\ref{fig:illusration} (c). 

However in practice, given a specific temperature, the optimization may not yield a point in the OZ, but rather a local optimal. We call the surrounding area of the local optimal with a comparable loss value a {\em false zone} (FZ). Since FZ does not include the GT, further search with decreasing temperatures must yield false results.
We observe that while deciding if an optimization result is a global optimal is impossible in {\em general optimization}, in our context (of backdoor defense), the problem is indeed tractable. According to Eq.~\ref{backdoor loss def}, the GT must have a very small loss regardless of the temperature. 
Therefore, first, we do not start reducing the temperature if the current result does not even have a small loss; second, even after we start to reduce the temperature, if we cannot achieve a small loss at a reduced temperature, we will first randomize the search, hoping to refocus to the OZ. If even the randomization does not yield a small loss, we backtrack to a previous (higher) temperature and resume search in a smoother space. The backtracking could be multi-step.

\smallskip
\noindent
{\em Example.} We use an example in Figure~\ref{fig:illusration} to illustrate our optimization. Assume the initial data point is $A$ in figure (a) (with $J$ the GT). The high temperature optimization in (a) yields $B$ which is a local optimal. However, its loss is very small (see its white color), comparable to the loss of the GT. 
Such a small loss grants 
temperature reduction and hence the search continues in the landscape in figure (b) (with a moderate temperature). It further narrows down to $C$.
Observe the zone of $C$ is white, denoting small loss values, and is part of the zone of $B$ in (a). The temperature further decreases and the search becomes in the landscape (c). It yields point $D$, which is a one-hot value but has a large loss (see its blueish color). The process backtracks to the previous temperature. With randomization, the search in (b) resumes from point $E$. The dashed edge from $D$ to $E$ denotes a random offsetting operation.
It then reaches $F$ in (b) 
and then $G$ in (c), which has a large loss as well. This time, our algorithm backtracks two steps as the last one-step backtrack yields a false result.  Therefore, the search roll-backs to the landscape in (a). With the random offset (from $G$ to $H$), the search is on track to reach the GT through the path $H\rightarrow I \rightarrow J$.

Our method can be formally defined as a constrained optimization problem: 
\begin{equation}
\label{Dynamic bound scaling}
\begin{aligned}
\small
\min \ &\lambda, \  s.t. \ \mathcal{L}_{inv} < \beta^\prime
\end{aligned}
\end{equation}
where $\lambda$ the temperature coefficient in Eq.~\ref{Softmax with t}  
and $\beta^\prime$ controls the loss boundary to allow temperature drop. 
At the beginning, $\lambda$ is high, the optimization is able to quickly converge to a loss value smaller than $\beta^\prime$. As $\lambda$ decreases, the landscape becomes rougher and the optimization becomes more difficult, then the loss may start increasing and $\lambda$ may be backtracked. 

The method is formally defined in Alg.~\ref{DBB}. 
Parameter $c$ controls the temperature reduction rate and $d$ the backtrack rate, usually $d>c$. In this paper, we use $d=5$ and $c=2$. We set the temperature upper bound $u = 2$ to avoid it grows too large. Parameter $\epsilon$ controls the random offset. 
Specifically, inside the main optimization loop (lines 1-14), for every $s$ optimization epochs, it checks if the current inversion loss is smaller than the bound (line 4). If so, the temperature is reduced (line 5). Otherwise, it backtracks (line 7) and applies random noise $\epsilon  \sim \mathcal{N}(0,\delta)$ to $w_{ij}$  with $\delta = 10$ (line 8).
Note that if the loss fails to go below $\beta'$ after $s$ epochs, the algorithm will backtrack more.

\begin{algorithm}[t]
\caption{Temperature Scaling and BackTracking}\label{DBB}
\hspace*{\algorithmicindent} \textbf{Input}: \ dataset $D^\prime$, target label $y$, params: $c,d,u,\epsilon,\beta^\prime$\\
\hspace*{\algorithmicindent} \textbf{Output}: invsered trigger $\alpha^*_{ij}$
\begin{algorithmic}[1]
\Repeat
\State Optimizing Eq.\ref{backddoor trigger inversion def relax}
\ForEach {$s$ epochs}
\If{$\mathcal{L}_{inv} < \beta^\prime$}
\State Temperature Focusing: $\lambda = \lambda / c$
\Else  
\State Backtracking:  $\lambda = \min(\lambda * d,u)$
\State Randomization:  $w_{ij} = w_{ij} + \epsilon, \epsilon \sim \mathcal{N}(0,\delta) $
\EndIf
\EndFor
\If{$\max{\alpha_{j} = 1, \forall i \in m}$ and $\mathcal{L}_{inv} < \beta^\prime$}
\State update best trigger: $\alpha^*_{ij} = \alpha_{ij}$
\EndIf
\Until {max epochs}

\end{algorithmic}
\end{algorithm}

\section{Evaluation}
\label{sec:eval}
\subsection{Experimental Setup}

\begin{figure}[t]
\vspace{-7pt}
\centering
\includegraphics[width=0.55\columnwidth]{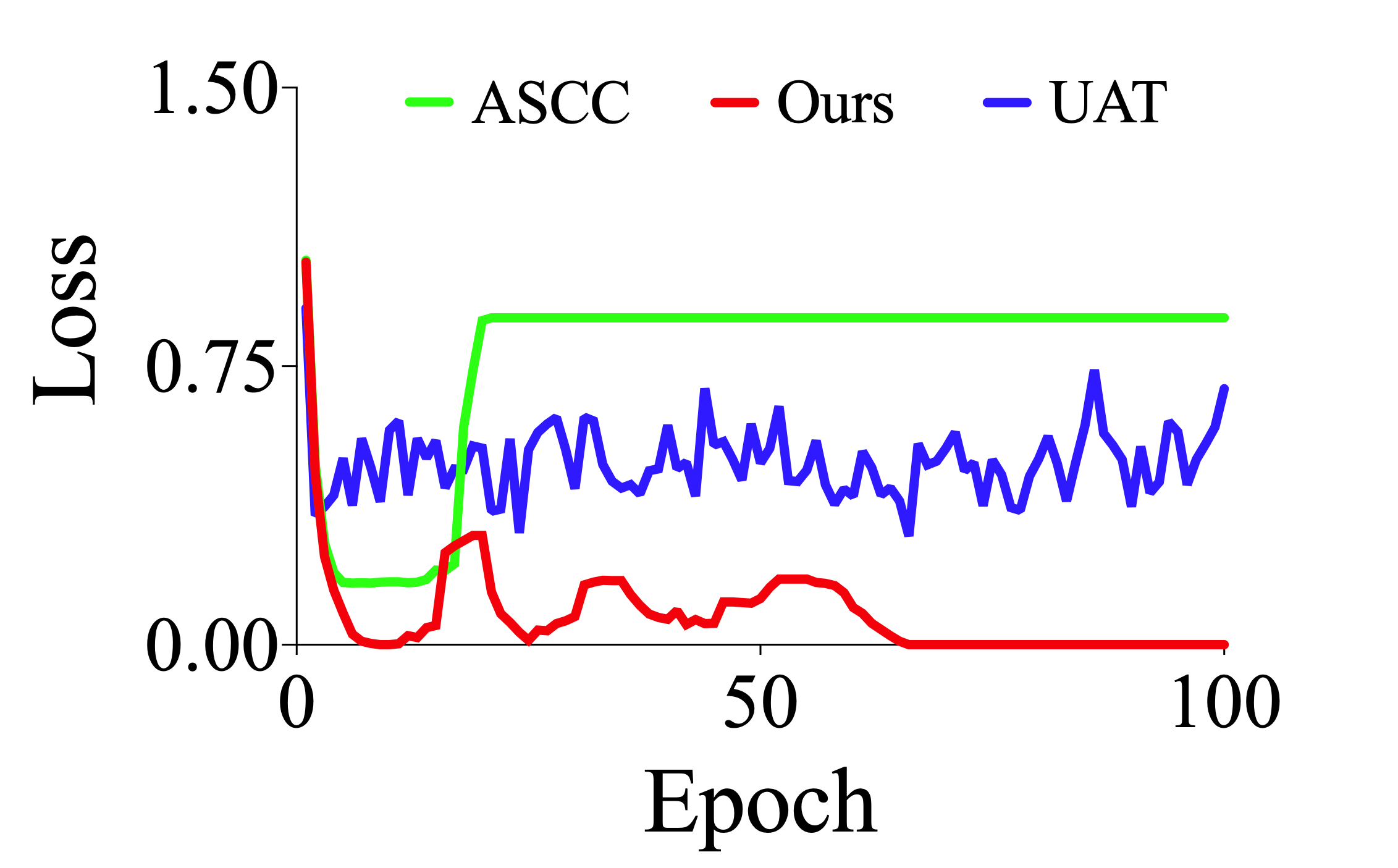}
\caption{Loss Comparison On A Trojaned Model (TrojAI \#R8-Test-ID-216)}
\vspace{-7pt}
\label{trojan model loss diff}
\end{figure}

\noindent
{\bf Tasks and Evaluation Models.}
We evaluate our method on 3 popular NLP tasks: Sentiment Analysis (SA)~\cite{socher2013recursive}, Name Entity Recognition (NER)~\cite{sang2003introduction,melamud2016context2vec} and Question Answering (QA)~\cite{rajpurkar2016squad}. We consider the 548 SA transformer models from the TrojAI competition~\cite{trojai} round 5, including 48 models from the training set and 480 models from the test set. Please see details in Appendix~\ref{app:settings}.
For each task, we leverage one fourth
of the  models in training set to derive the optimal threshold $\beta$ in Eq.~\ref{backdoor detection def rewrite}. 
For the (time-consuming) backdoor removal experiment, we randomly sample 20 trojaned models from each round to evaluate our method.
Since our method does not require training, we use both the training and test datasets in evaluation.

\begin{table*}[t!]
\footnotesize
\centering
\caption{Backdoor Detection Evaluation Results on TrojAI Datasets}
\label{trojai evaluation}
\scalebox{0.8}{
\begin{tabular}{llcccccccccccccccc}
\toprule
\multirow{2}{*}[-0.03in]{Task}  &\multirow{2}{*}[-0.03in]{Evaluation Set}  & \multicolumn{2}{c}{\textbf{Ours}} & \multicolumn{1}{c}{} & \multicolumn{2}{c}{ASCC}  & \multicolumn{1}{c}{} & \multicolumn{2}{c}{UAT}  & \multicolumn{1}{c}{} & \multicolumn{2}{c}{T-miner}  & \multicolumn{1}{c}{} & \multicolumn{2}{c}{GA} \\

\cmidrule{3-4} \cmidrule{6-7} \cmidrule{9-10} \cmidrule{12-13} \cmidrule{15-16}
 &     & \multicolumn{1}{c}{Acc} &    \multicolumn{1}{c}{Time(s)} & \multicolumn{1}{c}{} &  \multicolumn{1}{c}{Acc}  &  \multicolumn{1}{c}{Time(s)} & \multicolumn{1}{c}{}  &  \multicolumn{1}{c}{Acc}  &  \multicolumn{1}{c}{Time(s)} & \multicolumn{1}{c}{}  &  \multicolumn{1}{c}{Acc}  &   \multicolumn{1}{c}{Time(s)} & \multicolumn{1}{c}{} &  \multicolumn{1}{c}{Acc}  &   \multicolumn{1}{c}{Time(s)} \\

\midrule
\multirow{2}{*}{SA} &\multicolumn{1}{l}{R6 Train}           &\textbf{0.958} &\textbf{200}         &       &0.833  &199      &      &0.868 &256    &       &0.500  &3075    &        &0.848  &1666\\

\cmidrule{2-16}
    &\multicolumn{1}{l}{R6 Test}                           &\textbf{0.954} &\textbf{202}           &       &0.766  &209    &       &0.854  &251    &      &0.520   &3076   &       &0.798
    &1736\\
\midrule
\multirow{2}{*}{NER}    &\multicolumn{1}{l}{R7 Train}      &\textbf{0.917} &\textbf{580}           &       &0.738  &655    &       &0.889  &714    &       &-  &-         &   &0.700  &3675\\
\cmidrule{2-16}
 &\multicolumn{1}{l}{R7 Test}                              &\textbf{0.916} &\textbf{599}           &       &0.686  &640    &       &0.897  &718    &       &-  &-          &  &0.638  &3720 \\  
\midrule
\multirow{2}{*}{QA} &\multicolumn{1}{l}{R8 Train}          &\textbf{0.975}  &\textbf{530}          &       &0.808  &506    &       &0.892  &615    &       &-  &-          &   &0.767  &4281 \\
\cmidrule{2-16}
&\multicolumn{1}{l}{R8 Test}                               &\textbf{0.905}  &\textbf{532}          &       &0.695  &563    &       &0.783  &613    &       &-  &-          &   &0.667  &4310\\                        
\bottomrule

\end{tabular}
}
\vspace{-8pt}
\end{table*}

\begin{figure*}[t]
    \centering
    \subfigure[Continuous Optimization with No Constraint ]{
        \includegraphics[width=0.46\columnwidth]{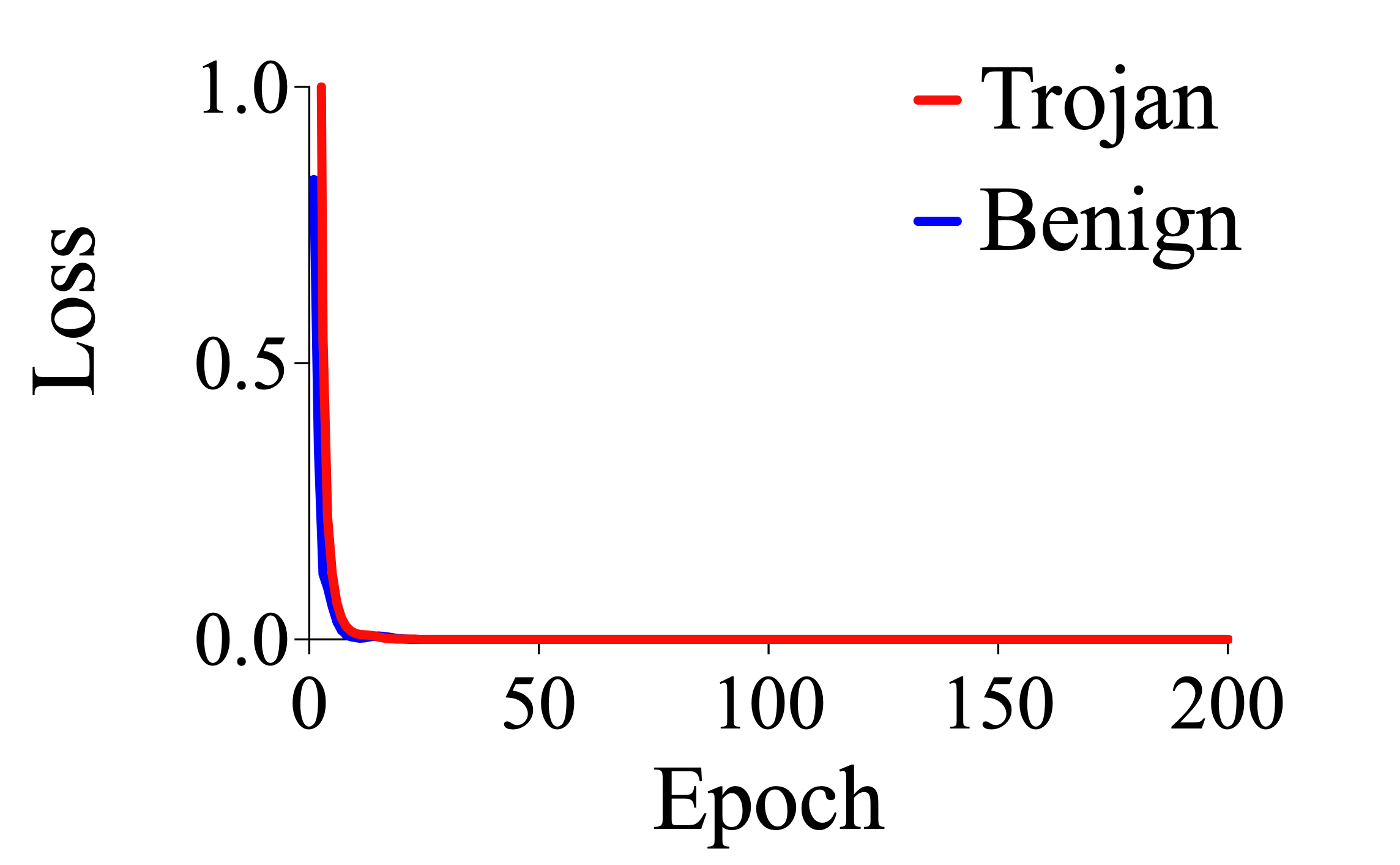}
        \label{continuous}
    }
    \subfigure[ASCC]{
	\includegraphics[width=0.46\columnwidth]{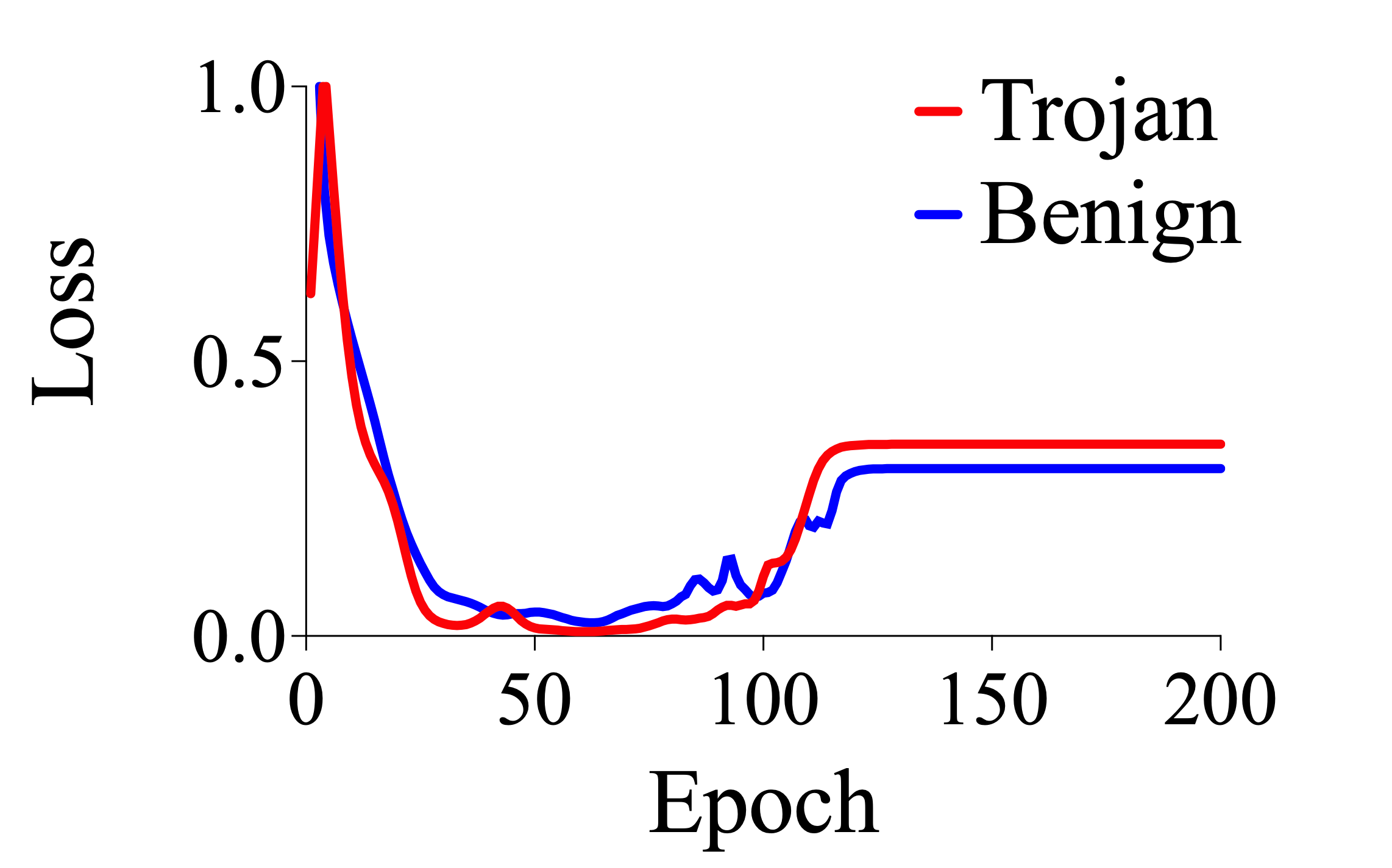}
	\label{ascc}
    }
    \quad
    \subfigure[UAT]{
	\includegraphics[width=0.46\columnwidth]{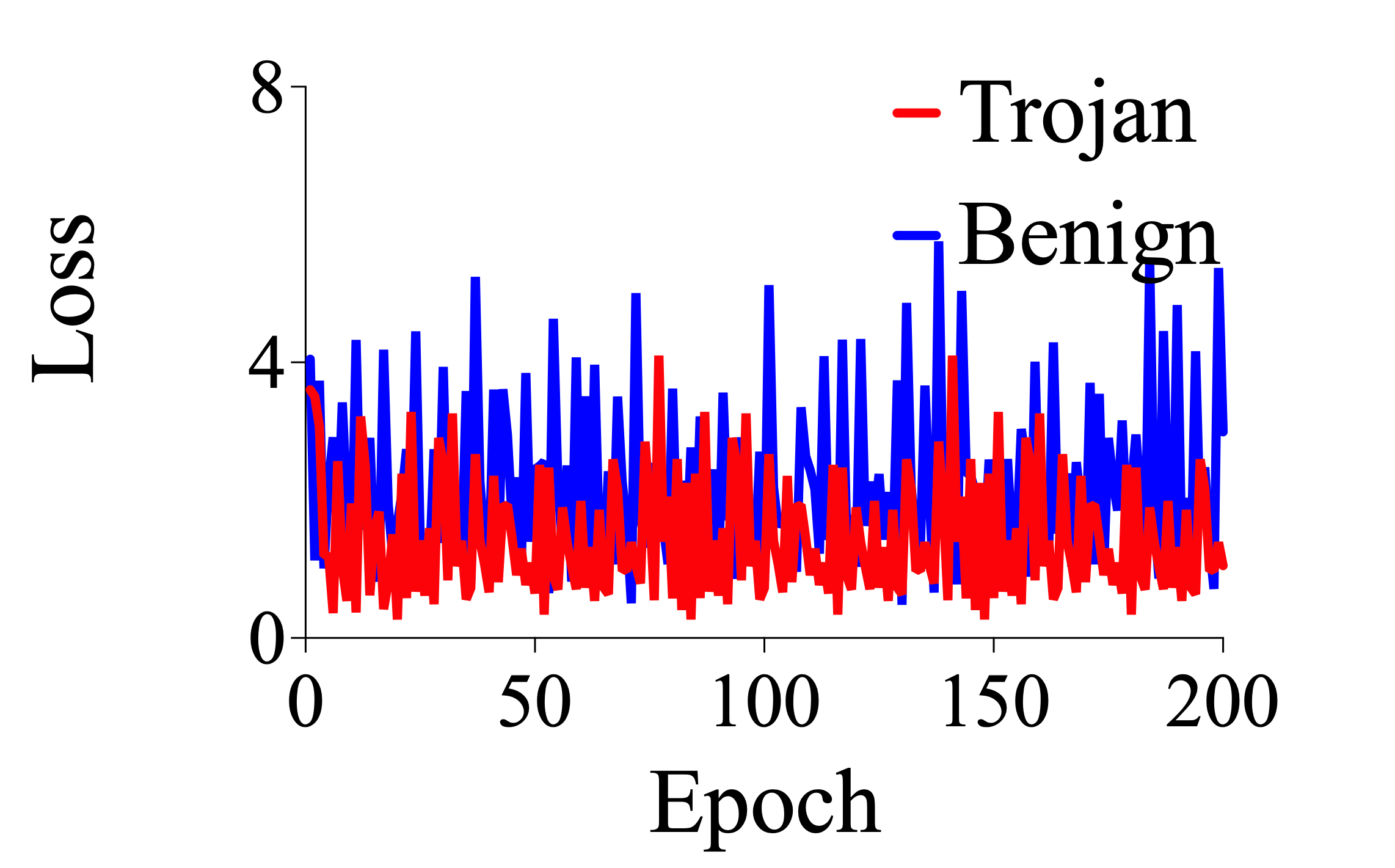}
	\label{uat}
    }
    \subfigure[Ours]{
	\includegraphics[width=0.46\columnwidth]{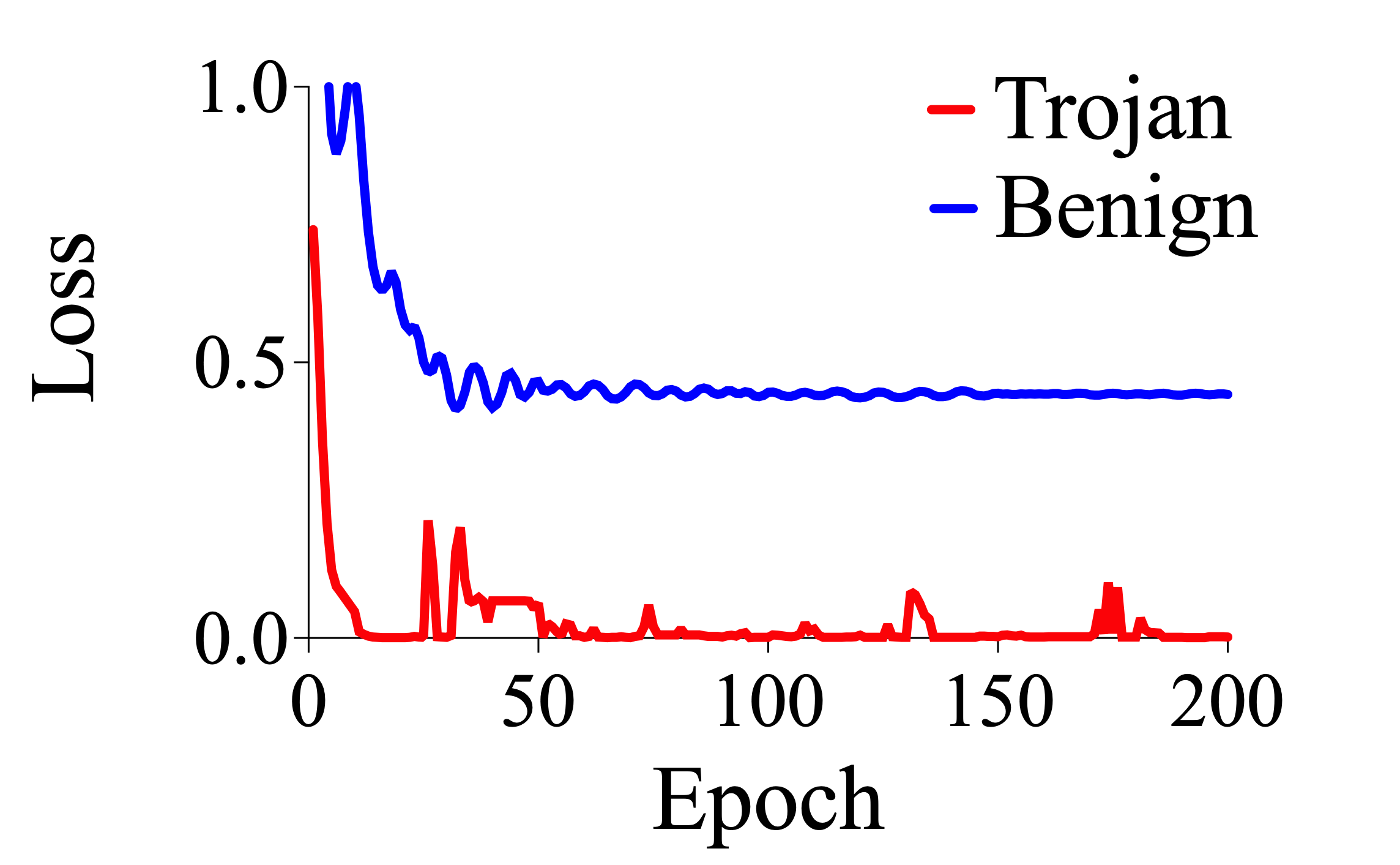}
	\label{ours}
    }
    \caption{Loss Comparison on Trojan/Benign model}
    \label{Trojan benign loss diff}
    \vspace{-5pt}
\end{figure*}


\noindent
{\bf Attack Setting.}
\textit{Standard Data Poisoning} (SDP)~\cite{gu2017badnets} is used for generating trojaned models in the TrojAI datasets. In particular, 9 poisoning configurations are used for the different tasks. Detailed settings can be found in Appendix~\ref{app:settings}.
In addition, we also evaluate  on the following 3 advanced NLP backdoor attacks.
\textit{Hidden Killer Attack}~\cite{qi2021hidden} proposes to use syntactic structure as trigger to achieve stealthiness.
\textit{Combination Lock Attack}~\cite{qi2021turn} rephrases sentences by substituting a set of words with their synonyms and  considers the substitution behaviors as the trigger. 
{\em SOS Attack}~\cite{yang2021rethinking} constructs a poisonous dataset by data augmentation to ensure the trigger sequence is only effective when all tokens appear in the input sentence.
For each attack, we use 20 trojaned and 20 benign models.
Specifically, hidden killer and combination lock models are trained on the SST-2 dataset with shared benign models~\cite{maas-EtAl:2011:ACL-HLT2011}, and the SOS models are trained on the IMDB movie review dataset~\cite{maas2011learning}. All codes are from original Github repos.

\smallskip
\noindent
{\bf Defense Setting.}
For backdoor detection, we use 20 samples in scanning, meaning that the trigger ought to flip majority of these samples.
For backdoor removal, we randomly select 20 trojaned models from the TrojAI training set in each round and first run our method to invert triggers, then optimize based on Eq.\ref{backdoor removal loss def rewrite} to unlearn the backdoors. Please see more details in Appendix~\ref{app:settings}.

\noindent
{\bf Baselines.}
We compare our method with 4 baselines for the backdoor detection task: ASCC~\cite{dong2021towards}, Genetic Algorithm (GA)~\cite{alzantot2018generating},  UAT~\cite{wallace2019universal}, and T-Miner~\cite{azizi2021t}. Their settings can be found in Appendix~\ref{app:settings}.

\begin{figure*}[t]
    \centering
    \subfigure[Hidden Killer Attack]{
        \includegraphics[width=2in]{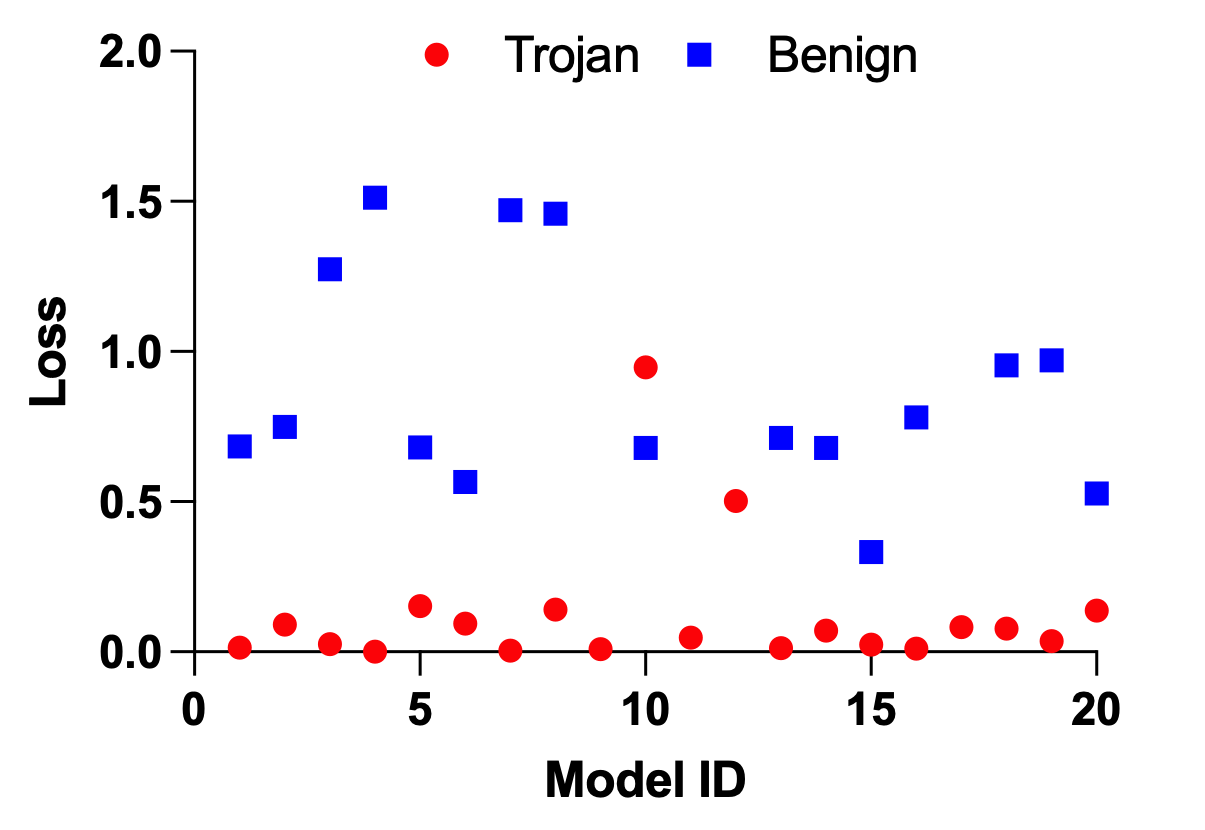}
        \label{hidden killer}
    }
    \subfigure[Combinational Lock Attack]{
	\includegraphics[width=2in]{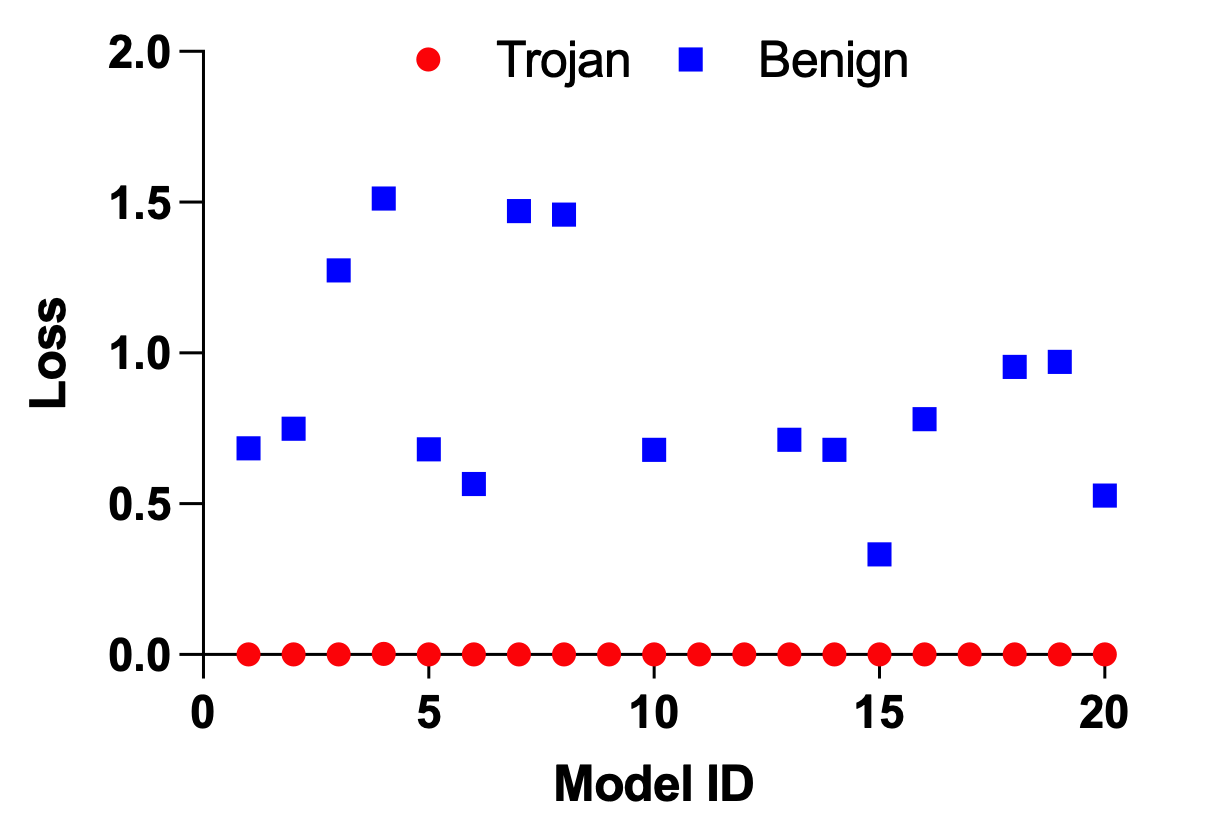}
	\label{bdk}
    }
    \subfigure[SOS Attack]{
	\includegraphics[width=2in]{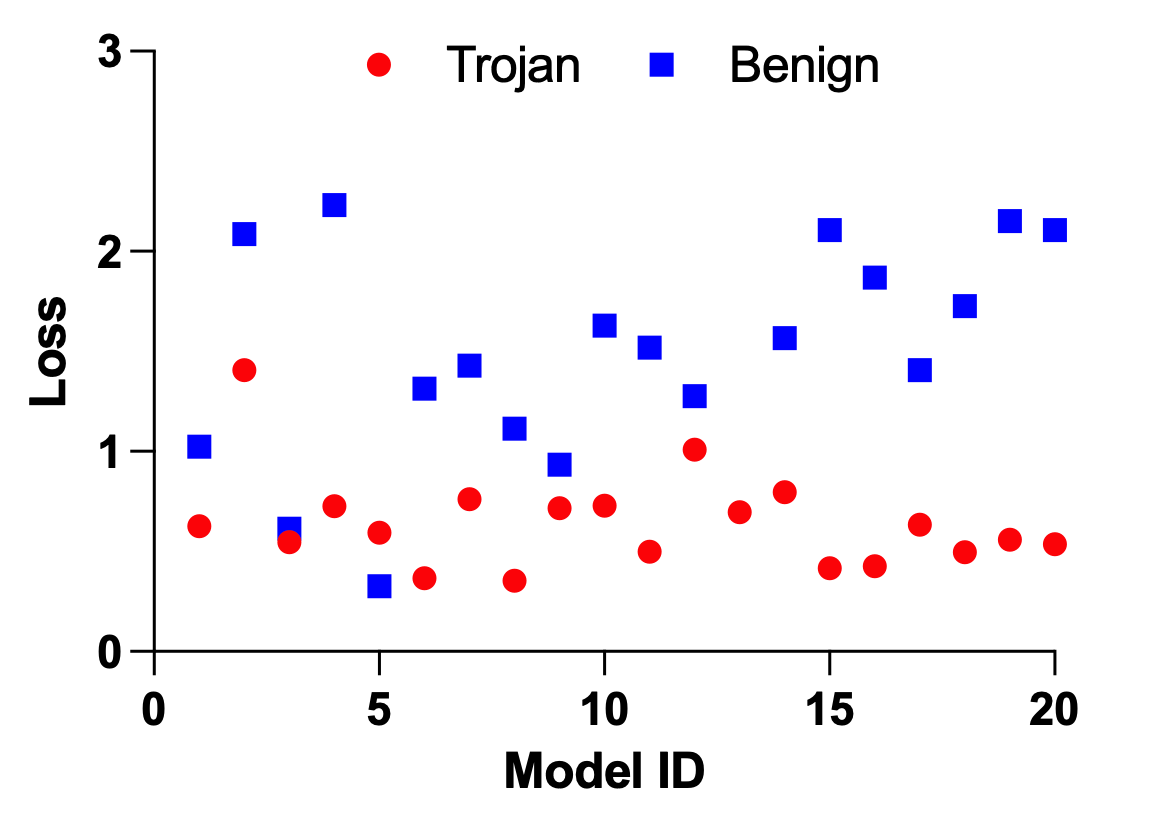}
	\label{sos}
    }
    \caption{Evaluation on Advanced NLP Backdoor Attacks}
    \label{Advanced Attack}
    \vspace{-7pt}
\end{figure*}

\subsection{Main Results}
\noindent
{\bf Backdoor Detection.}
We show the results on the 1548 TrojAI models in Table~\ref{trojai evaluation}. The first and second columns indicate the  tasks and the corresponding evaluation set.  In columns 3-12, every two columns show the detection accuracy and the time cost for a method. As shown in the third and fourth columns, our method is able to achieve over $90\%$ detection accuracy cross all tasks and outperforms all the baselines. For example, in the SA task, our method can achieve $0.958$ and $0.954$ on round 6 training and test sets, respectively. In contrast, the other two optimization based methods, ASCC and UAT, can only achieve $0.833$/$0.766$, and $0.868$/$0.854$, respectively. T-Miner achieves $0.5$ and $0.52$ on the round 6 training and test sets. Note that T-miner is a generator based detection method and may not scale well to large transformer models. GA can get $0.848$ and $0.798$ accuracy on round 6 which are comparable to ASCC. For the round 7 NER task, we find that UAT is more effective than other baselines, it can achieve $0.897$ on the test set, which is $21.1\%$ better than ASCC and $25.9\%$ better than GA. However, our method is still $1.9\%$ better. For the QA task, which is most challenging, our method has $0.905$ accuracy on the test set, which is $21\%$, $12.2\%$ and $23.8\%$ better than ASCC, UAT and GA, respectively. The time cost of our method is similar to the other two optimization methods due to the same setting, UAT is slightly slower than the other two due to its projection operation in each step. T-Miner and GA are more costly. In particular, T-Miner takes $14\times$ more computation time for the round 6 SA models when compared to our method. It is because T-Miner fine-tunes a generator model when scanning each subject model.  Our method is $8.6\times$, $6.21\times$ and $8.1\times$ faster than GA in the SA, NER and QA tasks.

Fig.~\ref{trojan model loss diff} and Fig.~\ref{Trojan benign loss diff} further explain the effectiveness of our method compared to the baselines. Fig.~\ref{trojan model loss diff} illustrates the loss value changes over the optimization epochs for different methods on a trojaned QA model from TrojAI round 8. The green line is for ASCC. As illustrated in the methodology section, it quickly falls into a large false zone (FZ), then the entropy regularization item forces the optimizer to find a one-hot value in the FZ. The value yields a large loss since it is not the ground truth (GT). The blue line is for UAT, it leverages first-order Taylor-approximation to derive the gradient direction of the current optimization variable and directly projects it to the discrete input space. In other words, every point in the  curve represents the loss of a one-hot token sequence. However, due to the rugged loss landscape, such local approximation might yield a wrong gradient direction, leading to the fluctuating curve and sub-optimal results. The red line is for our method.  In the early epochs, it also falls in some FZ as ASCC does. Later, due to  backtracking and randomization, it escapes the FZ's, reaches an optimal zone (OZ), and finally inverts the GT.  Fig.~\ref{Trojan benign loss diff} further compares the loss values between a trojaned and a benign SA models from TrojAI round 6 for each method. Due to the existence of FZ's, directly optimizing in the continuous space (with the default temperature) yields extremely small loss values for both trojaned and benign models as shown in Fig.~\ref{continuous}, leading to a large number of false positives in scanning. Fig.~\ref{ascc} and ~\ref{uat} illustrate the differences for ASCC and UAT, respectively. Observe that due to the ineffective search of the GT in these methods, the trojaned model's loss is not distinguishable from the benign model's, leading to  false negatives. In contrast, our method is able to achieve a clear separation between the benign and trojaned models as in Fig.~\ref{ours}.

We evaluate our method on the 3 advanced NLP backdoor attacks.
The results are in Fig.~\ref{Advanced Attack}. In each subplot,  we present the loss values of \textbf{optimal trigger estimations} by our method for individual models. The red cycles and blue boxes represent trojaned and benign models, respectively.  As in Fig.~\ref{hidden killer} and Fig.~\ref{bdk}, there are clear separations between trojaned and benign models for Hidden Killer and Combination Lock attacks. In particular, if we set $\beta=0.3$, we are able to achieve $0.95$ and $1.00$ detection accuracy for the two attacks. The SOS attack can be considered an adaptive attack for our method. It substantially reduces the size of OZ via adversarial data augmentation, which makes it difficult for the optimizer to find the GT. As shown in Fig.~\ref{sos}, although the red and blue separation becomes much smaller than the other two attacks, we are still able to get $0.875$ if we set the $\beta=1$. 


\begin{table}[t!]
\footnotesize
\centering
\caption{Backdoor Removal Results on TrojAI Datasets}
\label{trojai removal evaluation}
\scalebox{0.8}{
\begin{tabular}{lccccccc}
\toprule
\multirow{2}{*}[-0.03in]{Task} & \multicolumn{2}{c}{Before Removal} & \multicolumn{1}{c}{} & \multicolumn{2}{c}{After Removal} \\


 
\cmidrule{2-3} \cmidrule{5-6} 
&   \multicolumn{1}{c}{Clean Acc} &    \multicolumn{1}{c}{ASR} & \multicolumn{1}{c}{} &   \multicolumn{1}{c}{Clean Acc} &   \multicolumn{1}{c}{ASR}\\
\midrule 
\multicolumn{1}{c}{SA}  &0.909   &0.980   &   &\textbf{0.892} &\textbf{0.051}     \\
\midrule
\multicolumn{1}{c}{NER}  &0.978   &0.957   &   &\textbf{0.959} &\textbf{0.032}     \\
\midrule
\multicolumn{1}{c}{QA}  &0.727   &0.998   &   &\textbf{0.720} &\textbf{0.091}     \\
\bottomrule

\end{tabular}
}
\vspace{-6pt}
\end{table}

\begin{table}[t!]
    \footnotesize
    \centering
    \caption{Adaptive Attack}
    \label{adaptive attack}
    \tabcolsep=4pt
    \scalebox{0.8}{
        \begin{tabular}{lccccccccccc}
        \toprule
        \multirow{2}{*}[-0.03in]{Parameter} & \multicolumn{2}{c}{SA} & & \multicolumn{2}{c}{NER}  & & \multicolumn{2}{c}{QA} \\

        
         
        \cmidrule{2-3} \cmidrule{5-6} \cmidrule{8-9} 
        & \multicolumn{1}{c}{Acc} & \multicolumn{1}{c}{ASR} & &   \multicolumn{1}{c}{Acc} & \multicolumn{1}{c}{ASR} & &   \multicolumn{1}{c}{Acc} & \multicolumn{1}{c}{ASR}\\
        \midrule

        $\phi=0$ & 0.950 & 0.991 & & 0.900 & 0.941 & & 0.900 & 0.999 \\
        $\phi=0.1$ & 0.900 & 0.959 & & 0.900 & 0.942 & & 0.900 &0.962  \\
        $\phi=0.5$ & 0.850 & 0.853 & & 0.850 & 0.812 & & 0.800 & 0.794 \\
        $\boldsymbol \phi \boldsymbol = \boldsymbol 1$ & \textbf{0.650} & \textbf{0.531} & & \textbf{0.550} & \textbf{0.448} & & \textbf{0.600} & \textbf{0.428} \\
        \bottomrule
        \end{tabular}
    }
\end{table}

\noindent
{\bf Backdoor Removal.}
We show the backdoor removal results on 60 randomly selected models for the 3 NLP tasks across 3 TrojAI rounds (20 for each round) in Table\ref{trojai removal evaluation}. The first column presents the task. The second and third columns show the trojaned models' clean accuracy and attack success rate (ASR) on the test set before removal. The fourth and fifth columns show the average clean accuracy and ASR after removal. Observe that for all tasks, our method is able to reduce the ASR down to less than $10\%$ with slight clean accuracy degradation. 
Note that there are a number of highly effective backdoor removal methods in the visiom domain~\cite{wang2019neural,wu2021adversarial,li2021neural,tao2022model,liu2018fine}. These techniques cannot be easily adapted to the NLP domain which has different characteristics, such as sparse space, discrete pipeline, and sequential input. 

\subsubsection{Ablation Study}
We study the contributions of our design choices on a subset of TrojAI models, with 20 trojaned and 20 benign for each task. The results show that the optimization has low precision scores without temperature scaling/reduction and low recall scores without backtracking. It renders the importance of both design choices. Please see details in Appendix~\ref{app:ablation}.

\subsubsection{Adaptive Attack}
Besides the SOS attack, we devise another adaptive attack which is targeting our Assumption \uppercase\expandafter{\romannumeral2}. The idea is to encourage samples stamped with triggers not to have a extremely small loss. In particular, we revise the target loss in Eq.\ref{backdoor loss def} as follows. 
\begin{equation}
\label{adaptive backdoor loss def}
\begin{aligned}
\small
\mathcal{L}{\mathcal{D}_p}(\theta^*) &= \mathop{\mathbb{E}}_{(x,y)\sim  \mathcal{D}} \mathcal{L} (f(x;\theta^*),y) \\
 &+ \mathop{\mathbb{E}}_{(x^*,y^*) \sim  \mathcal{D}^*} \max(\mathcal{L} (f(x^*;\theta^*),y^*) - \phi, 0)
\end{aligned}
\end{equation}
where $\phi$ is a hyper-parameter to control the trigger loss during poisoning. For different $\phi$ values, we train 10 trojaned models and mixed them with the same number of benign models for each task. According to Table~\ref{adaptive attack}, our method stays effective when $\phi=0.5$, the detection accuracy is over $80\%$ for all tasks. As $\phi$ further increases, the method becomes less effective. However, the ASR drops a lot as well, which means the attack is not effective anymore. Intuitively, the ASR is entangled with the training loss. Therefore, it's difficult for attackers to achieve both high ASR and moderate loss value to bypass our method.





\section{Conclusions}
We present a novel optimization method for NLP model backdoor defense. It reduces the trigger inversion problem to inverting weight value vectors for all tokens in the dictionary. It then dynamically scales the temperature in the {\it softmax} function to gradually narrow down the search space for the optimizer, helping it to produce a close-to one-hot inversion result with high ASR. The one-hot value hence corresponds to the ground truth trigger. It also utilizes temperature backtracking to escape local optimals. Our experiments show that the method is highly effective and produces better backdoor scanning and removal results than the baselines.

\bibliography{example_paper}
\bibliographystyle{icml2022}

\clearpage
\newpage
\appendix

\section*{Appendix}
\label{appendix}

\begin{table*}[ht]
\footnotesize
\centering
\caption{Ablation Study}
\label{ablation}
\scalebox{1}{
\begin{tabular}{ccccccccccccccc}
\toprule
\multirow{2}{*}[-0.03in]{Method} & \multicolumn{3}{c}{QA} & \multicolumn{1}{c}{} & \multicolumn{3}{c}{NER}  & \multicolumn{1}{c}{} & \multicolumn{3}{c}{QA} \\


 
\cmidrule{2-4} \cmidrule{6-8} \cmidrule{10-12} 
&   \multicolumn{1}{c}{Precision} &    \multicolumn{1}{c}{Recall} &    \multicolumn{1}{c}{Acc} & \multicolumn{1}{c}{} &   \multicolumn{1}{c}{Precision} &    \multicolumn{1}{c}{Recall} &    \multicolumn{1}{c}{Acc} & \multicolumn{1}{c}{} &   \multicolumn{1}{c}{Precision} &    \multicolumn{1}{c}{Recall} &    \multicolumn{1}{c}{Acc}\\
\midrule

\multicolumn{1}{c}{w/o Temp Scaling}  &0.559   &0.950   &0.600 &   &0.581   &0.900   &0.625 & &0.563   &0.900   &0.600     \\
\midrule
\multicolumn{1}{c}{w/o Backtracking} &1.000   &0.800   &0.900 &   &0.936   &0.750   &0.850  & &1.000   &0.700   &0.850     \\
\midrule
\multicolumn{1}{c}{\textbf{Ours}}  &\textbf{1.000}   &\textbf{0.950}   &\textbf{0.975} &   &\textbf{0.947}   &\textbf{0.900}   &\textbf{0.925} & &\textbf{1.000}   &\textbf{0.850}   &\textbf{0.925}     \\
\bottomrule

\end{tabular}
}
\end{table*}

\section{Detailed Experimental Settings}
\label{app:settings}

\noindent
{\bf Tasks and Evaluation Models.}
We evaluate our method on 3 popular NLP tasks: Sentiment Analysis (SA)~\cite{socher2013recursive}, Name Entity Recognition (NER)~\cite{sang2003introduction,melamud2016context2vec} and Question Answering (QA)~\cite{rajpurkar2016squad}. The SA models are trained on 7 different datasets from Amazon review~\cite{ni2019justifying} and IMDB~\cite{maas-EtAl:2011:ACL-HLT2011} to output binary predictions (i.e., positive and negative).  For NER, we consider the 540 TrojAI round 7 models, in which 180 from the training set and 360 from the test set. The datasets used to train these NER models include CoNLL-2002~\cite{10.3115/1119176.1119195} with 4 name entities, the BNN corpus~\cite{weischedel2005bbn} with 4 name entities and OntoNotes~\cite{hovy-etal-2006-ontonotes} with 6 name entities. A model is supposed to identify all the valid entities in given samples.   For the QA task,  we evaluate the 120 and 360 models from the TrojAI round 8 training and test sets, respectively. The QA models are trained on 2 public datasets: SQUAD V2~\cite{2016arXiv160605250R} and SubjQA~\cite{bjerva20subjqa}. Given a pair of question and context, a QA model is supposed to predict the position of answer in the context. In total, there are 7 transformer architectures: DistilBERT~\cite{Sanh2019DistilBERTAD}, GPT-2~\cite{radford2019language}, BERT~\cite{bert}, RoBERTa~\cite{roberta}, MobileBERT~\cite{sun2020mobilebert}, Deepset~\cite{deepset} and Google Electra~\cite{clark2020electra}.

\noindent
{\bf Attack Setting.}
\textit{Standard Data Poisoning} (SDP)~\cite{gu2017badnets} is used for generating trojaned models in the TrojAI datasets. In particular, 9 poisoning configurations are used for the different tasks. Specifically, based on the attack goal, there are two types of attacks: {\em universal attack} which flips samples from all classes to the target class and {\em label specific attack} which only flips samples from a certain victim class to the target class. Based on the trigger effect range, there are {\em global trigger}, which is effective at any position in the input sentence, {\em local trigger} which can only cause misclassification when inserted at a certain position (range). For example, in the SA task, a local trigger can be set as effective only when it is in the first half of an input sentence. In the QA task, a sample local trigger can trigger misclassification only if it is stamped in the question. A trigger can be a single character, word or a short phrase.

\noindent
{\bf Defense Setting.}
For backdoor detection, we use 20 samples per class in scanning, meaning that the trigger ought to flip majority of these samples. 
For SA, we use an auxiliary benign model during the optimization to insure that the inverted words are neutral (to avoid false positives). The benign model is randomly drawn from the TrojAI model set. 
For the NER models, as the number of labels is relatively large, we apply a strategy similar to K-Arm~\cite{shen2021backdoor}, in which we run 20 epochs for each victim-target label pair and select the most promising top two pairs based on the loss value for further optimization. 
The 
same strategy is applied for all the optimization related baselines. 

For backdoor removal, we randomly select 20 trojaned models from the TrojAI training set in each round and first run our method to invert triggers, then optimize based on Eq.\ref{backdoor removal loss def rewrite} to unlearn the backdoors. We use $10\%$ of the training data and $20\%$ of the chosen training samples are stamped with our inverted trigger and marked with the correct labels. The clean accuracy and ASR before and after removal are evaluated on the whole test set. 

\noindent
{\bf Baselines.}
We compare our method with 4 baselines for the backdoor detection task: ASCC~\cite{dong2021towards}, Genetic Algorithm (GA)~\cite{alzantot2018generating},  UAT~\cite{wallace2019universal}, and T-Miner~\cite{azizi2021t}.
ASCC~\cite{dong2021towards} and GA~\cite{alzantot2018generating} were proposed to generate NLP adversarial examples. They can be easily adapted for backdoor trigger inversion. In particular, we change their search space from small synonym lists to the entire dictionary.  For ASCC, we set the coefficient of the sparsity regularization term to 10 as suggested in the paper. For GA, we set the population size to 300 and the mutation rate to 0.5 and the number of generations to 10. For UAT, we set $k$=1 in the top-k token search. ASCC and UAT are evaluated on all the models in the TrojAI datasets. For the other two costly methods, we evaluate GA on 248 SA models, 120 NER models and 180 QA models with half trojaned and half benign, and T-Miner only on the 248 SA models from TrojAI round 6.  Note that T-Miner was proposed for the SA task and requires substantial efforts to extend it to other tasks. 

\noindent
{\bf Parameters Setting.}
We set the length of trigger to 10 (i.e., inverting 10 weight vectors).
We set the number of optimization epochs to 200 and use the Adam~\cite{kingma2014adam} optimizer with the initial learning rate 0.5. All optimization related baseline methods share the same configuration. All experiments are done on a machine with a single 24GB memory NVIDIA Quadro RTX 6000 GPU.

\section{Ablation Study}
\label{app:ablation}
We study the contributions of our design choices on a subset of TrojAI models, with 20 trojaned and 20 benign for each task. As shown in Table~\ref{ablation}, without temperature scaling/reduction, the optimization keeps searching in the continuous space and converges inside false zones,  
leading to the low precision scores and hence the low accuracy. 
Without backtracking, the algorithm continues to reduce the temperature and eventually makes the optimizer walk in a very rugged (close to discrete) space. As a result, it produces many false negatives, leading to the low recall scores and hence the low accuracy.

\end{document}